# Application of Dimensional Reduction in Artificial Neural Networks to Improve Emergency Department Triage During Chemical Mass Casualty Incidents

Nicholas D. Boltin[1], Joan M. Culley[2], Homayoun Valafar[1*]

[1]Department of Computer Science and Engineering, University of South Carolina, Columbia, SC

[2]College of Nursing, University of South Carolina, Columbia, SC

*Corresponding Author: homayoun@cse.sc.edu

**ABSTRACT**

*Chemical Mass Casualty Incidents (MCI) place a heavy burden on hospital staff and resources. Machine Learning (ML) tools can provide efficient decision support to caregivers. However, ML models require large volumes of data for the most accurate results, which is typically not feasible in the chaotic nature of a chemical MCI. This study examines the application of four statistical dimension reduction techniques: Random Selection, Covariance/Variance, Pearson's Linear Correlation, and Principle Component Analysis to reduce a dataset of 311 hazardous chemicals and 79 related signs and symptoms (SSx). An Artificial Neural Network pipeline was developed to create comparative models. Results show that the number of signs and symptoms needed to determine a chemical culprit can be reduced to nearly 40 SSx without losing significant model accuracy. Evidence also suggests that the application of dimension reduction methods can improve ANN model performance accuracy.*

## INTRODUCTION & BACKGROUND

Improving patient wait times and length of stay in Emergency Departments (ED) has the ability to improve patient quality of care and reduce hospital and emergency response costs. Studies have shown that increasing a patient's length of stay by as much as two hours can cost the hospital more than $3 million annually. Likewise, ED crowding is associated with inferior health care and loss of revenue[1-4]. Previous work has shown that this problem can potentially be remedied by the use of some optimization techniques[5, 6]. However, hospitals have always been burdened with collecting as much information as possible while efficiently triaging all patients with accurate precision. Healthcare providers are now looking to utilize modern technology to assist caregivers with complex decision-making.

WISER[7] is a software decision support system developed by the National Library of Medicine (NLM) and is designed to assist emergency responders in hazardous material incidents. It provides a wide range of information on hazardous substances, including substance identification support, physical characteristics, human health information, and containment/suppression advice. Its key features include rapid access to essential information about a hazardous substance via NLM's Hazardous Substance Data Bank (HSDB), containing detailed peer-reviewed information on hazardous substances and comprehensive decision support.

Previous work done by this lab has demonstrated that chemical identification accuracy could be improved by integrating machine learning algorithms into WISER's substance ID support tool[8, 9]. This study aims to continue improving WISER's support system by reducing the number of signs and symptoms (SSx) needed to identify a hazardous chemical through statistical dimension reduction techniques. By reducing the number of SSx needed, we can reduce the time required to evaluate a patient. This will increase triage efficiency while maintaining information integrity and reduce the time patients wait to see a caregiver. Ultimately, a more efficient triage will reduce the length of stay and improve patient quality of care.

## METHODS

*Description of the Training and Testing Data-sets*

The dataset used for training artificial neural networks in this study was collected by reviewing the toxicology information in WISER, which is derived from NLM's Hazardous Substance Data Bank (HSDB). A listing of 438 chemicals containing 79 associated signs and symptoms was created. An example of the resultant table is shown in Figure 1. Each of the 438 substances found in WISER is represented in the first column, where columns 2-80 represent the corresponding 79 SSx found in WISER for a given chemical. Each row represents a chemical's SSx profile, where binary values are assigned according to the presence or absence (0 or 1, respectively) of each SSx.

Examination of the created dataset revealed several substances with identical SSx profiles. In instances where chemicals contained the same profile, this cluster of chemicals was reduced to a single observation. It is understood that two chemicals that produce the same signs and symptoms may not be treated in the same way regarding patient care. However, reducing these sub-groups to a single representation was necessary to remove any bias towards a particular chemical profile. After removing the duplicated observations, the original listing of 438 substances was reduced to 311(shown in Figure 1) uniquely distinguishable chemicals, serving as the reverse-engineered list of unique chemicals.

Figure 1. Section of WISER's Reconstructed Database. NLM's toxicology information stored in the Hazardous Substance Data Bank (HSDB) was used to verify and reverse engineer signs and symptoms associated with each chemical.

Signs and symptoms related to victims of an actual chemical incident would be ideal for testing the newly trained ANN models. However, accurate patient records during mass casualty incidents are limited and usually incomplete [10, 11]. Therefore, three additional synthetic data sets were created to test the model's performance after training. To precisely control the amount of missing or inaccurate data, simulated patients with SSx profiles were generated from the ideal dataset of 311 unique substances by perturbation of randomly selected SSx. Each substance was replicated 100 times to create a reasonably extensive test set of 31,100 simulated patients. Three test sets were generated by randomly toggling SSx at a 5%, 10%, and 15% selection rate. Probability density profiles shown in Figure 2 were examined to ensure random selections of perturbed SSx across each simulated patient test set.

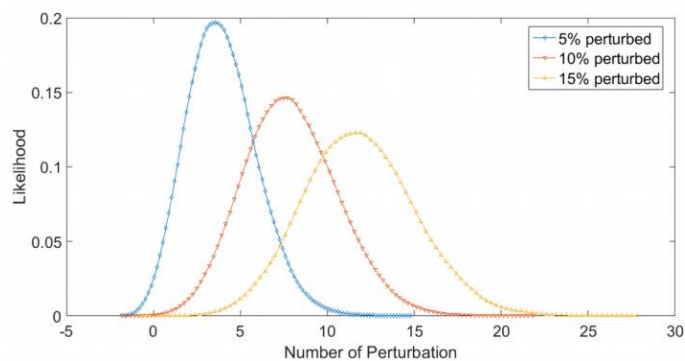

Figure 2. The Kernel Density Estimation of the three additional test sets. Test datasets were created by starting with the ideal dataset of 311 unique substances from WISER and changing the presence of chemical symptoms by 5%, 10%, and 15%.

*Design, Training, and Testing of Artificial Neural Networks*

A systematic pipeline was developed to create and optimize the Artificial Neural Network (ANN) models. Models were built in the Matlab 2016Ra environment using the pattern recognition toolbox. The model used in this study was a scaled conjugate gradient backpropagation Artificial Neural Network[12] and is illustrated in Figure 3. Our process began by importing the dataset of 311 unique substances with their 79 SSx profiles. Each input node of the input layer represents an individual chemical SSx profile. ANN models were created using a standard 70/30 split, where 70% of the data was used to train the ANN, and 30% was used to test the ANN. The output error was then calculated on the ANN's ability to classify chemicals in the 30% test set. However, due to the unique nature of the chemical profiles it was necessary to resample each observation to optimize the ANN model. Resampling increased the number of observations from 311 to 1,555, with each chemical being replicated 5 times. Replication also insured that each unique chemical profile would be represented in the Training and Testing data sets. Figure 4 shows that by increasing the dataset by five iterations, the prediction accuracy increases from <1% to 99%. Additional testing was performed on the ANN using the artificially created patient test sets described earlier. Using addition test sets with increasing degrees of error in the profiles is comparable to collecting patient SSx's during a chemical incident was used to measure the ANN model's robustness to false or missing information

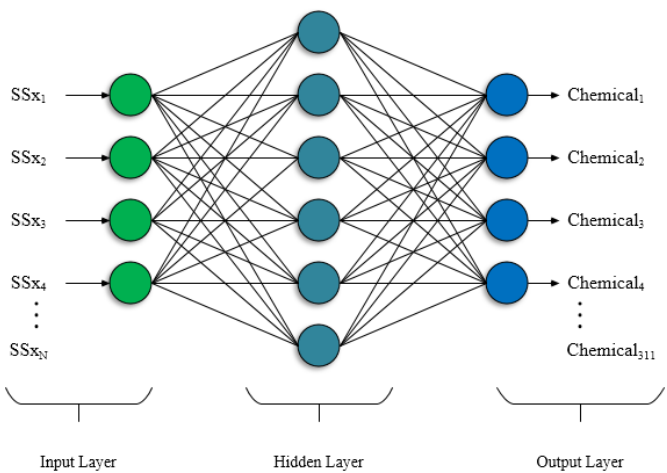

Figure 3. Scaled conjugate gradient backpropagation ANN model for chemical classification based on signs and symptoms found in the NLM's Hazardous Substance Database.

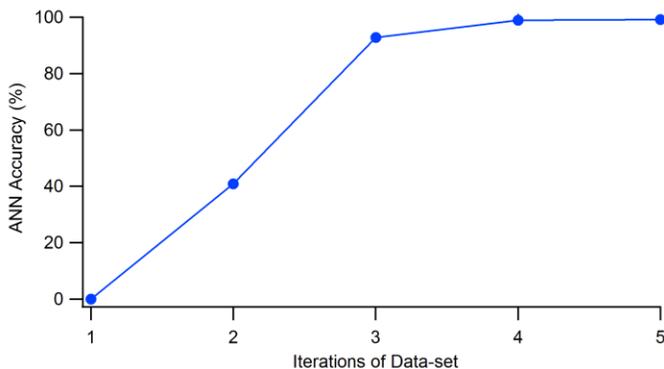

Figure 4. Optimizing the ANN model prediction accuracy by resampling each chemical profile. The number of iterations represent the number of times the Chemical SSx profile was replicated. The increase in accuracy can be attributed to each chemical being represented in the training and testing data sets.

*Dimension Reduction*

Creating an ANN model trained to classify 311 chemicals allowed us to measure the accuracy of deploying such models when caregivers can collect all 79 SSx. We recognized that in some scenarios, such as mass casualty incidents (MCIs), it might not be feasible to obtain and develop a complete patient profile. If all 79 SSx cannot be collected, it may be practical to use dimension reduction techniques to reduce the amount of SSx necessary to classify a chemical and maintain a degree of accuracy in the ANN model. To explore reducing the number of SSx, we have utilized the following popular statistical methods for dimension reduction, random feature selection, variance/covariance, correlation coefficients, and principal component analysis. To measure the performance of these methods, each dimension reduction technique was used to determine 40 SSx. These 40 SSx were then used to create ANN models similar to the 79 SSx model in the previous section. In addition, because an ANN model's accuracy is dependent on the number of hidden layers used during training, we tested models at increments of 10 hidden layers, starting at 10 and ending at 100 hidden layers. The average performance accuracy was measured at each increment of hidden layers.

*All 79 SSx*

One hundred ANN models were created using all original 79 SSx found in the chemical dataset. This was done to set a standard for which future models would be compared. Ten models were trained to start at 10 hidden neurons to obtain an average accuracy and then sequentially increased by steps of 10 hidden networks. This allowed us to calculate the overall performance accuracy of the model and determine the number of hidden networks to use that would maximize the model's efficiency. Additional testing was also performed on the model using the test-sets perturbated at 5%, 10%, and 15%.

*First 40 SSx (Alphabetically)*

The first method used to reduce the number of SSx needed to predict a chemical from 79 to 40 SSx was choosing random symptoms. For simplicity, SSx were ordered alphabetically, and the first 40 SSx were chosen, which significantly reduced the dataset to nearly half the original size. To compare the results, 100 ANN models were created, and average performance accuracy was calculated using the same method described for the 79 SSx.

*40 SSx based on Covariance/Variance*

ANN models were then created by reducing the original dataset from 79 SSx to 40 SSx based on the variation between SSx. For any two random SSx vectors A and B, the covariance between A and B can be described using Equation 1, where N is the number of observations, $\mu_A$ is the mean of A, $\mu_B$ is the mean of B, and * denotes the complex conjugate. A 79x79 covariance matrix was created by a pairwise covariance calculation between each SSx column observations in the original dataset. The diagonal vector of the covariance matrix describes the variation of the 79 SSx and can be defined by Equation 2, where μ is the mean of A and defined by Equation 3. The binary distribution of the data made it unnecessary to normalize the dataset. The workflow pipeline was then followed to create 100 ANN models and calculate their prediction accuracy.

$$\frac{1}{N-1}\sum_{i=1}^{N}(A_i - \mu_A)*(B_i - \mu_B) \qquad \text{Equation 1}$$

$$V = \frac{1}{N-1}\sum_{i=1}^{N}|A_i - \mu|^2 \qquad \text{Equation 2}$$

$$\mu = \frac{1}{N}\sum_{i=1}^{N} A_i \qquad \text{Equation 3}$$

### 40 SSx based on Correlation Coefficient

Pearson's linear correlation coefficients were calculated for each of the 79 SSx in the original unique chemical dataset and can be defined by Equation 4, where $X_a$ is one of the columns in the original data matrix X, and $n$ is the length of each column. Correlation coefficients range between -1 and +1 where a value of -1 indicates a perfect anti-correlation between the SSx, while a value of +1 indicates a perfect positive correlation between the SSx. An example of the correlation vector for Arrhythmia can be seen in Figure 5. The figure shows that Tachycardia, Bradycardia and Hypotension Shock are positively correlated indicating that when the symptom Arrhythmia is present, there is a strong likelihood that Tachycardia, Bradycardia or Hypotension Shock will be present as well. A correlation coefficient threshold of 0.4555+/- was found to reduce the original dataset from 79 SSx to 40 SSx with the least similarity. These 40 uncorrelated SSx were used to create one hundred ANN models using the designed workflow and test the model's accuracy to predict harmful chemicals.

$$\bar{X}_a = \sum_{i=1}^{n} X_{a,j}/n \qquad \text{Equation 4}$$

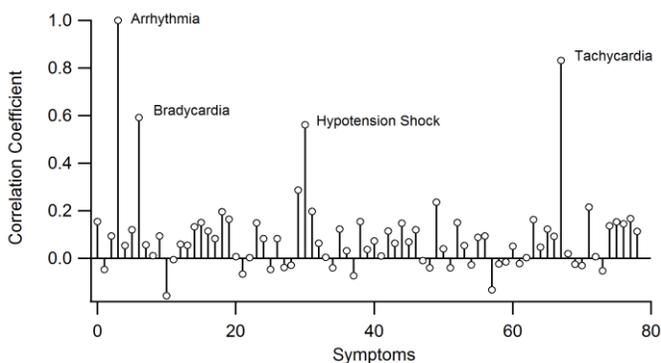

Figure 5. The Correlation Coefficient Vector for Arrhythmia. The plot shows that bradycardia, hypotension shock, and tachycardia have a strong positive correlation with Arrhythmia.

### 40 PCs based on Principal Component Analysis

The original 311x79 dataset was analyzed using principal component analysis. Single value decomposition was used to evaluate the 79 SSx and can be defined by Equation 4.5, where $U$ is the orthonormal matrix with the eigenvectors of $XX^T$, $S$ is the diagonal matrix with the singular values, and $V^T$ is the orthonormal matrix with the eigenvectors of $X^TX$. 40 principal components were selected by selecting the first 40 columns of $X$. The proportion of variance explained in each principal component was calculated by squaring the standard deviation of each Principle Component (PC) and then dividing by the trace or total sum of variance. The first 40 PCs were used to create 100 ANN models using the same protocol as used for previous techniques described and prediction accuracies were calculated.

$$X_{mn} = U_{mm}S_{mn}V_{nn}^T \qquad \text{Equation 5}$$

## RESULTS

The average performance accuracy was calculated for each of the dimension reduction techniques described in the previous section. Table 1 describes the overall average performance for each of the dimension reduction techniques when training the ANN models and performing additional testing using the 5%, 10% and 15% perturbated datasets. Table 2 describes the number of hidden networks used where the model's prediction accuracy experienced the best performance. Figure 8 shows the model prediction accuracy for each of the techniques used and is discussed further in the following sections.

### All 79 SSx

ANN models were created using all 79 SSx associated with the chemical dataset. Figure 8A shows the model prediction accuracy when training the ANN and for all additional test-sets. When training the ANN models using all 79 SSx, the average performance accuracy was 99.5% across all hidden networks. When testing the ANN models with the 5%, 10%, and 15% datasets, the performance accuracy was 73.3%, 43.5%, and 23.9% respectively. The model performed best at 100 hidden networks with 99.9% accuracy, and when additional testing was done on the model using the 5%, 10%, and 15% perturbated datasets, the model performed with 83%, 51%, and 27% accuracy respectively.

### First 40 SSx (Alphabetically)

ANN models were created using the first 40 SSx chosen alphabetically as inputs. Figure 8B shows the model prediction accuracy when training with only the first 40 SSx and for testing with all additional test-sets. The average performance for training ANN models with 40 alphabetic SSx was 97.1%. Additional testing with 5%, 10% and 15% data-sets demonstrated an overall accuracy of 65.2%, 38.1% and 21.2% respectively. The model's best training performance was at 80 hidden networks with 2.7% error and an accuracy of 97.2%. When tested with the 5%, 10% and 15% perturbated datasets the model performed with 71%, 42%, and 23% accuracy respectively.

### 40 SSx based on Covariance/Variance

Figure 6 shows the distribution of variance for the 79 SSx in the chemical dataset. The 40 SSx were selected by examining the highest values in the variance vector. ANN models were created using the 40 SSx with the largest variation in their data as inputs. Figure 8C shows the model prediction accuracy when training with 40 SSx based on variance and for testing with all additional test-sets. When training the ANN models, the average accuracy for all hidden networks was 98.1% for identifying chemicals in the original test-set. For chemicals in the 5%, 10%, and 15% perturbated test-sets, the ANN model's performance accuracy was 72.6%, 46.2%, and 27.2% respectively. The ANN model performance was best at 40 hidden networks with 98.5% accuracy and when doing additional testing using the 5%, 10% and 15% perturbated datasets the model performed with 79%, 52%, and 32% accuracy respectively.

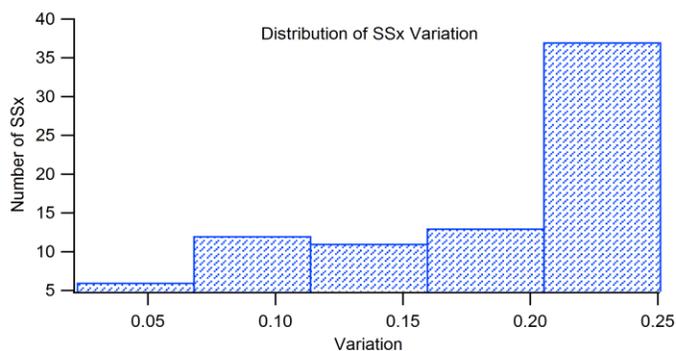

Figure 6. The distribution of symptom variance for the chemical data set. 79 SSx was reduced to 40 SSx by selecting the SSx which had the highest variation values.

*40 SSx based on Correlation Coefficient*

ANN models were created using the 40 SSx with the least amount of correlation in the chemical dataset. Figure 8D shows the ANN model prediction accuracy when training with 40 SSx based on correlation and for testing with all additional test-sets while optimizing the hidden networks from ten nodes to one hundred nodes. The overall performance accuracy for training ANN models based on the 40 uncorrelated SSx was 98.8%. The overall performance of ANN models tested with the 5%, 10%, and 15% test-sets was 65.6%, 38.9%, and 21% respectively. The best-trained model saw a performance error of 1.07% and an accuracy of 98.9% at 70 hidden networks. When additional testing was done using the 5%, 10%, and 15% perturbated test-sets, the ANN model performed with 73%, 44%, and 25% accuracy respectively.

*40 PCs based on Principal Component Analysis*

The scree plot seen in Figure 7A shows that the first principal component (PC), which accounts for the most variability in the dataset, explains 8.8% of the variability in the original dataset or variance. The second PC explains 6.9% of the total variance. Figure 7B shows the cumulative variance for the original dataset of all 79 SSx. It was determined that the first 40 PCs cumulatively explains 89% of the variability in the original data. The top 40 principal components were used to create one hundred ANN models. Figure 8E shows the ANN model prediction accuracy when training with 40 PCs and for testing with all additional test-sets while optimizing the hidden networks from ten nodes to one hundred nodes. Training ANN models with 40 principal components produced an overall prediction accuracy of 99.8%. With additional testing using the 5%, 10% and 15% perturbated test-set the overall prediction accuracy was 74.1%, 46.1% and 25.7% respectively. The ANN model performance was best at 60 hidden networks with 99.9% accuracy and when doing additional testing using the 5%, 10% and 15% perturbated test-sets the ANN model performed with 81%, 52%, and 29% accuracy respectively.

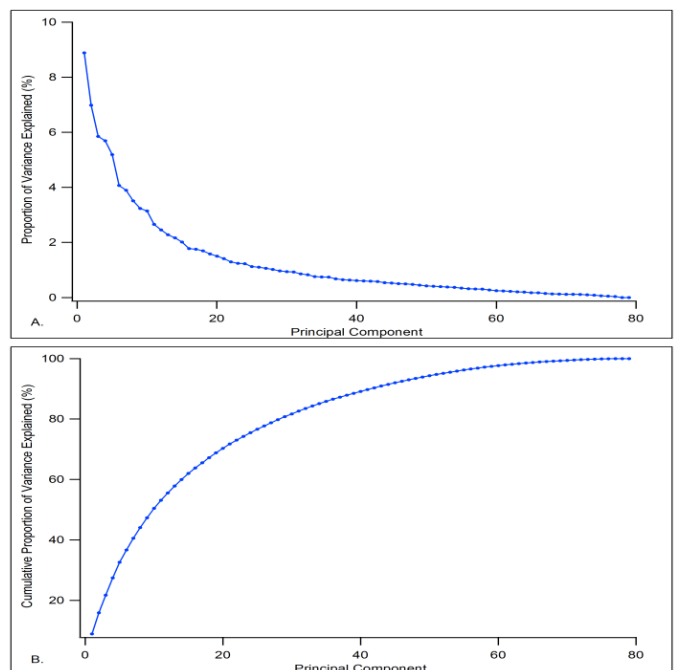

Figure 7. PCA Variation. (A) Proportion of variance that each principal component explains. (B) The cumulative summation of each principal component. The first 40 principal components explain ~89% of the variability in the original dataset

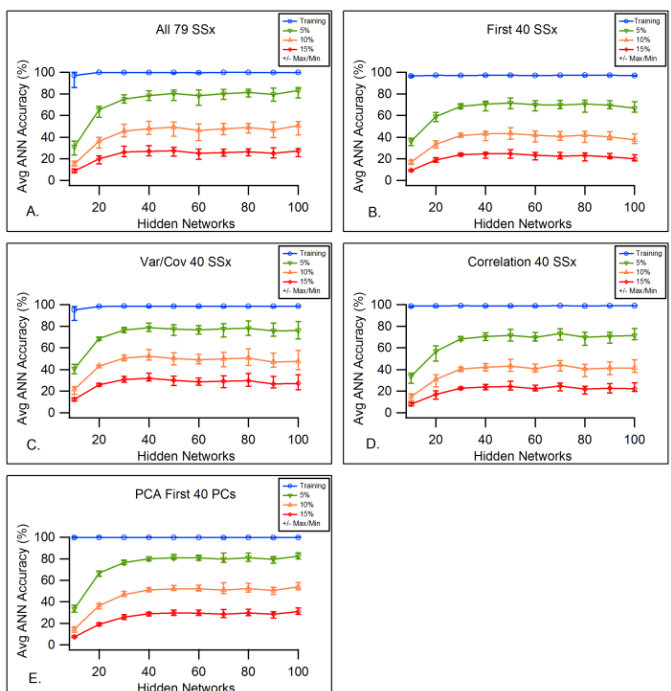

Figure 8. Prediction accuracy for ANN models using (A) All 79 SSx, (B) Frist 40 random SSx, (C) 40 SSx based on highest variation, (D) 40 SSx based on least correlation, (E) First 40 principal components. The average prediction accuracy was calculated for models with the training dataset and the 5%, 10% and 15% perturbated datasets.

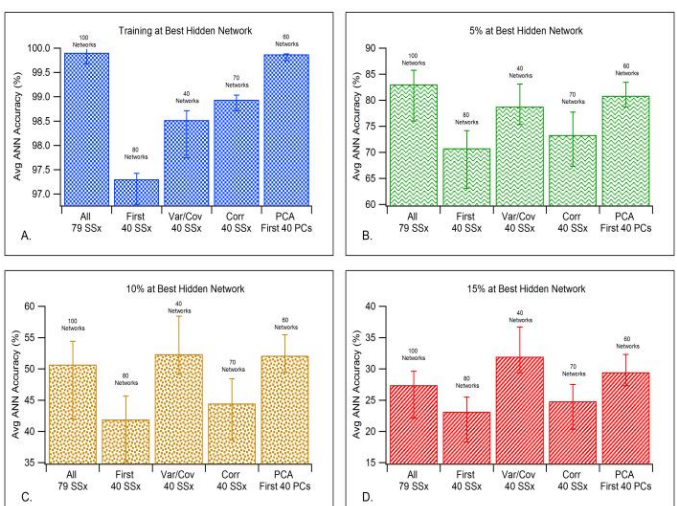

Figure 9. Comparing ANN Model Prediction Accuracy. (A) Comparison of ANN model performance during training. (B) Comparison of ANN model performance when tested with the 5% perturbated test-set. (C) Comparison of ANN model performance when tested with the 10% perturbated test-set. (D) Comparison of ANN model performance when tested with the 15% perturbated test-set.

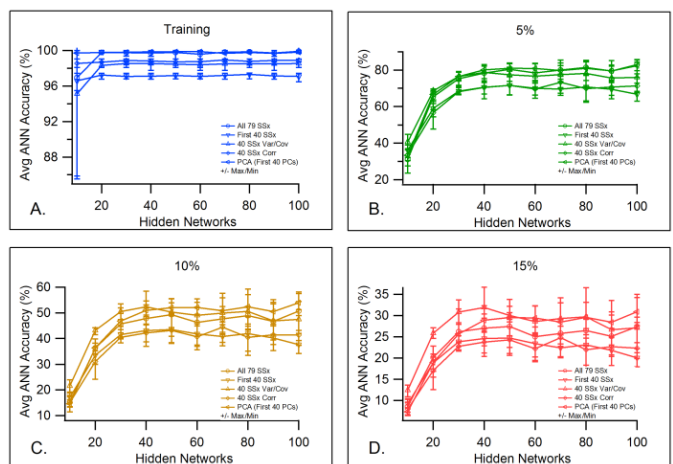

Figure 10. Comparing ANN Models with Different Test-set. (A) Average ANN model prediction accuracy for all dimensional reduction techniques during training. (B) Average ANN model prediction accuracy for all dimensional reduction techniques when tested with the 5% perturbated test-set. (C) Average ANN model prediction accuracy for all dimensional reduction techniques when tested with the 10% perturbated test-set. (D) Average ANN model prediction accuracy for all dimensional reduction techniques when tested with the 15% perturbated test-set.

## DISCUSSION

Table 1 describes the overall performance accuracy of each of the dimension reduction techniques. When training the ANN models, all DRTs were able to classify chemicals with a high degree of accuracy with PCA performing the best overall at 99.8% and the first 40 alphabetical SSx performing the worst at 97.1%. This would make sense, seeing as the first 40 alphabetical SSx were not chosen based on any correlation or variation what-so-ever. For this reason, there could be SSx in the selected dataset with high correlation which would provide little additional information to the model or SSx with a high degree of variation that is missing from the dataset which could have provided the model with valuable decision-making information. Table 1 also shows that when we start to introduce inaccuracies in the data the ANN model's overall performance will diminish. When SSx were perturbated by 5%, which is the equivalent of changing approximately four SSx from their correct value to an incorrect value, each of the DRT's accuracy was reduced by an average of 28%. This is most notably seen in ANN models create using the first 40 alphabetic SSx and models created based on a correlation threshold where performance accuracy dropped to 65.2% and 65.6% respectively.

Table 1. The Overall Average Performance Accuracy for ANN models created using all 79 signs/symptoms and for each of the dimension reduction techniques.

| Model | Average ANN Performance Accuracy (%) | | | |
| --- | --- | --- | --- | --- |
| | Training | 5% | 10% | 15% |
| All 79 SSx | 99.5 | 73.3 | 43.5 | 23.9 |
| First 40 SSx | 97.1 | 65.2 | 38.1 | 21.2 |
| 40 SSx Var/Cov | 98.1 | 72.6 | 46.2 | 27.2 |
| 40 SSx Corr | 98.8 | 65.6 | 38.9 | 21.0 |
| PCA (First 40) | 99.8 | 74.1 | 46.1 | 25.7 |

If we assume that ANN models trained with all 79 SSx to be the standard, meaning that we would not expect any models created using a reduction technique to perform better, we can then compare DRT models to the 79 SSx ANN performance. Examining the best performance accuracy, seen in Figure 9 for each model allows us to compare ANN models trained with DRT and the 79 SSx ANN standard. In Table 2 we see that models trained with 40 principal components performed the same as the standard while all others performed at least a degree less. When we performed additional testing using the 10% and 15% perturbated datasets we see that models create with PCA and models created using variance performed better than the 79 SSx standard. This may suggest that selecting precise SSx based on their variation and information gain may be more robust than just adding SSx that would provide little additional information and will even reduce the accuracy in predicting chemicals.

Table 2. Best Number of Hidden Networks. Average prediction accuracy of ANN models created with dimension reduction techniques at the hidden networks that had the best performance.

| Model | Best# of Hidden Networks | Average ANN Performance Accuracy (%) | | | |
|---|---|---|---|---|---|
| | | Training | 5% | 10% | 15% |
| All 79 SSx | 100 | 99.9 | 83.0 | 50.7 | 27.4 |
| First 40 SSx | 80 | 97.3 | 70.8 | 41.9 | 23.1 |
| Cov/Var 40 SSx | 40 | 98.5 | 78.7 | 52.4 | 31.9 |
| Corr 40 SSx | 70 | 98.9 | 73.3 | 44.4 | 24.8 |
| PCA 40 PCs | 60 | 99.9 | 80.8 | 52.1 | 29.4 |

## CONCLUSIONS

In general, this work has demonstrated that utilizing dimension reduction techniques can be an effective way of determining the sign and symptoms necessary to make a chemical classification. When we compare the ANN models created with DRTs to the standard model which used all 79 SSx, we see that each of the models performed similarly during training and with the additional testing. With an optimized number of hidden networks, ANN models trained with 40 SSx can outperform 79 SSx and show greater robustness to inaccurate data. This work demonstrates that artificial neural networks can be used to improve decision support tools used to give guidance to chemical exposures such as WISER and that collecting 40 SSx can be just as effective as collecting 79 SSx.

## ACKNOWLEDGMENTS

Research reported in this publication was supported by a grant from the National Institutes of Health (5R01LM011648 and P20 RR-01646100)